\documentclass[authoryear,12pt,review]{elsarticle}

\usepackage{lineno}
\usepackage{setspace}
\modulolinenumbers[1]

\journal{Geography Compass}

\usepackage{natbib}
\usepackage{amsmath}
\usepackage{mathabx}
\usepackage{booktabs}
\usepackage{balance}
\usepackage{color,soul}
\usepackage{subcaption}
\usepackage{float}
\usepackage[margin=1in]{geometry}
\usepackage{url}

\usepackage{tabularx}

\makeatletter
\newcommand*{\centerfloat}{%
  \parindent \z@
  \leftskip \z@ \@plus 1fil \@minus \textwidth
  \rightskip\leftskip
  \parfillskip \z@skip}
\makeatother

\newcommand\blfootnote[1]{%
	\begingroup
	\renewcommand\thefootnote{}\footnote{#1}%
	\addtocounter{footnote}{-1}%
	\endgroup
}

\begin{document}

\pagenumbering{gobble}

\begin{frontmatter}

\title{Geo-Text Data and Data-Driven Geospatial Semantics}
%
\author{Yingjie Hu}
\ead{yjhu.geo@gmail.com}
%
%
%
\address{GSDA Lab, Department of Geography, University of Tennessee, Knoxville, TN, 37996, USA}

\begin{abstract}
Many datasets nowadays contain links between geographic locations and natural language texts. These links can be geotags, such as geotagged tweets or geotagged Wikipedia pages, in which location coordinates are explicitly attached to texts. These links can also be place mentions, such as those in news articles, travel blogs, or historical archives, in which texts are implicitly connected to the mentioned places. This kind of data is referred to as \textit{geo-text data}. The availability of large amounts of geo-text data brings both challenges and opportunities. On the one hand, it is challenging to automatically process this kind of data due to the unstructured texts and the complex spatial footprints of some places. On the other hand, geo-text data offers unique research opportunities through the rich information contained in texts and the special links between texts and geography. As a result, geo-text data facilitates various studies  especially those in data-driven geospatial semantics. This paper  discusses geo-text data and related concepts. With a focus on data-driven research, this paper systematically reviews a large number of studies that have discovered multiple types of knowledge from geo-text data. Based on the literature review, a generalized workflow is extracted and key challenges for future work are discussed. 
\end{abstract}


\end{frontmatter}


\openup 0.1em

\section{Introduction}
Recent years have witnessed an unprecedented increase in the volume, variety, and velocity of data from different sources \citep{miller2015data}. Thanks to the advancements in sensors and information technologies, authoritative organizations, such as the U.S. Geological Survey, are continuing to produce many datasets with often richer content and higher precision. Meanwhile, general individuals, with the support of GPS-enabled smart devices, are also contributing large amounts of data via social media platforms, online blogs, review websites, and others \citep{goodchild2007citizens,haklay2008web}. As a result, various types of datasets have been generated.   \blfootnote{\textit{How to cite this article:} Hu Y. Geo-text data and data-driven geospatial semantics. Geography Compass. 2018, e12404. \url{https://doi.org/10.1111/gec3.12404}}

Among these datasets, there is one kind that contains interesting links between geographic locations and natural language texts. Some of these links are geotags, such as geotagged tweets or geotagged Wikipedia pages, in which location coordinates are directly attached to texts. Some other links are in the form of place mentions, such as those in news articles or travel blogs, in which the texts mention one or multiple place names. This kind of data is referred to as \textit{geo-text data} in this paper. 

To a certain degree, geo-text data already exists in many GIS applications. In a vector crime map, the locations of crimes can be linked to crime types, such as ``Property Crime" and ``Violent Crime", which are represented as text strings. In a raster land-use map, pixels representing geographic locations are linked to land-use categories, such as ``Commercial" and ``Agricultural". Although these crime types and land-use categories are texts, they are pre-defined by a schema and can only be chosen from a set of pre-defined text strings. This type of texts with a pre-defined schema is considered as \textit{structured data}. By contrast, the texts in geo-text data (e.g., geotagged tweets) can be composed freely, and should be considered as \textit{unstructured data}. Structured data can be handled relatively easily, since they can be converted to numeric indices based on the related schema. It is more challenging to process unstructured data due to language flexibility and text ambiguity. Despite these challenges, natural language texts offer rich information, such as  keywords, topics, entities, and sentiments, and enable various new research when linked to geographic locations. 

Geo-text data has been used in many empirical studies. 
For example, geotagged tweets were employed for supporting disaster response \citep{huang2015geographic,zhang2015towards} and investigating public health issues \citep{widener2014using,stefanidis2017zika}. Geotagged Flickr photos were used to enrich gazetteers \citep{kessler2009bottom} and to represent vague places \citep{hollenstein2010exploring}. Natural language  descriptions about landscapes were collected to understand the perceptions of people and their sense of place \citep{mark2011landscape,wartmann2018describing}.
Meanwhile, methods and tools were developed in geographic information retrieval (GIR) for recognizing and geo-locating place names from texts \citep{jones2008geographical}. Examples of such tools, also called \textit{geoparsers}, include MetaCarta \citep{frank2006spatially}, GeoTxt \citep{karimzadeh2013geotxt}, Edinburgh Geoparser \citep{alex2015adapting}, and TopoCluster \citep{delozier2015gazetteer}. Open and labeled datasets, such as the corpora annotated using SpatialML \citep{mani2010spatialml}, WikToR \citep{gritta2017s}, and GeoCorpora \citep{doi:10.1080/13658816.2017.1368523}, were made available for training and testing geoparsers. While many empirical studies exist,  there lacks a systematic discussion on the concept of geo-text data, its formal representation, and the types of knowledge that can be extracted.
This paper fills such a gap by bringing together the insights from different studies and organizing them using a coherent framework. Specifically, this paper makes the following contributions:
\begin{itemize}
\item A formal representation of geo-text data based on a general GIS theory.
\item A systematic discussion on the knowledge that can be extracted from geo-text data.
\item A generalized workflow for processing and analyzing geo-text data.
\item A set of key challenges for future research based on geo-text data.
\end{itemize} 

The remainder of this paper is organized using a series of questions. Section 2  addresses the question ``\textit{what is geo-text data?}" by presenting a formalization of geo-text data and discussing  related concepts. Section 3 examines the question ``\textit{what can we get from geo-text data?}" by reviewing a large number of studies that have discovered various types of knowledge from geo-text data. Section 4 answers the question ``\textit{what is a general workflow that we can follow to analyze geo-text data?}" by presenting a workflow generalized from the literature. Section 5 addresses the question ``\textit{what are some key challenges for future research?}". Finally, Section 6 summarizes this work.

\section{Geo-text data}
\textit{Geo-text data} can be considered as a collective term that encompasses many specific types of data, such as geotagged social media, geotagged Wikipedia pages, news articles, historical archives, location-focused online reviews, geotagged housing posts, and others that contain links between locations and texts. While these different types of data have been used in separate studies, they share similar characteristics and can be processed and analyzed using similar methods. This paper abstracts from the specific studies and data formats, and identifies the core concepts and methods that can be applied to geo-text data in general. The identified knowledge could be integrated to educational programs on \textit{spatial data science}. 

Based on a general theory of geographic representation in GIS \citep{goodchild2007towards}, geo-text data can be formalized as Equation \ref{formalization}.
\begin{equation}
\langle f,[t],W \rangle    \label{formalization}
\end{equation} 
where:
\begin{itemize}
	\item $f$ is the geographic location, or \textit{spatial footprint}, of geo-text data. A common spatial footprint is a point defined by a pair of coordinates, such as the location of a geotagged tweet. However, $f$ can also be lines (e.g., roads and rivers), polygons (e.g., cities and states), and polyhedras (e.g., a 3D geographic feature). In addition, $f$ can be vague places, cognitive regions, or other forms that lack crisp boundaries \citep{montello2003s,jones2008modelling,montello2014vague}.
	
	\item $t$ is the timestamp associated with the data record. $t$ is optional and therefore is surrounded by the square brackets in Equation \ref{formalization}. Many geo-text data have timestamps, such as the posting time of a tweet, the publishing time of a news article, and the editing time of a Wikipedia page. Such time information can enable valuable time series analysis. 
	
	\item  $W$ is the textual content of geo-text data. $W$ can be modeled using a simple bag-of-words model, while more sophisticated methods, such as parse trees \citep{schuster2016enhanced}, can be employed to provide more accurate modeling of texts. Depending on the specific needs of an application, different types of information can be extracted from $W$, such as entities (e.g., persons, places, and events), topics (e.g., music and travel), and sentiments (e.g., happy and sad).   
\end{itemize}

Compared with typical GIS data, such as temperature measurements and digital elevation models (DEM), in which numeric values are attached to locations, geo-text data can have  a  sentence, a paragraph, or even an entire article attached to a location.  Geo-text data also includes Flickr photo tags \citep{hollenstein2010exploring,tardy2016semantic}, geotagged surnames of people \citep{longley2011creating,cheshire2012identifying}, street or mountain names \citep{hill2000core,alderman2016place}, and others, in which words and phrases, instead of complete sentences, are attached to locations.

\begin{figure}[h]
	\centering
	\includegraphics[width=0.95\textwidth]{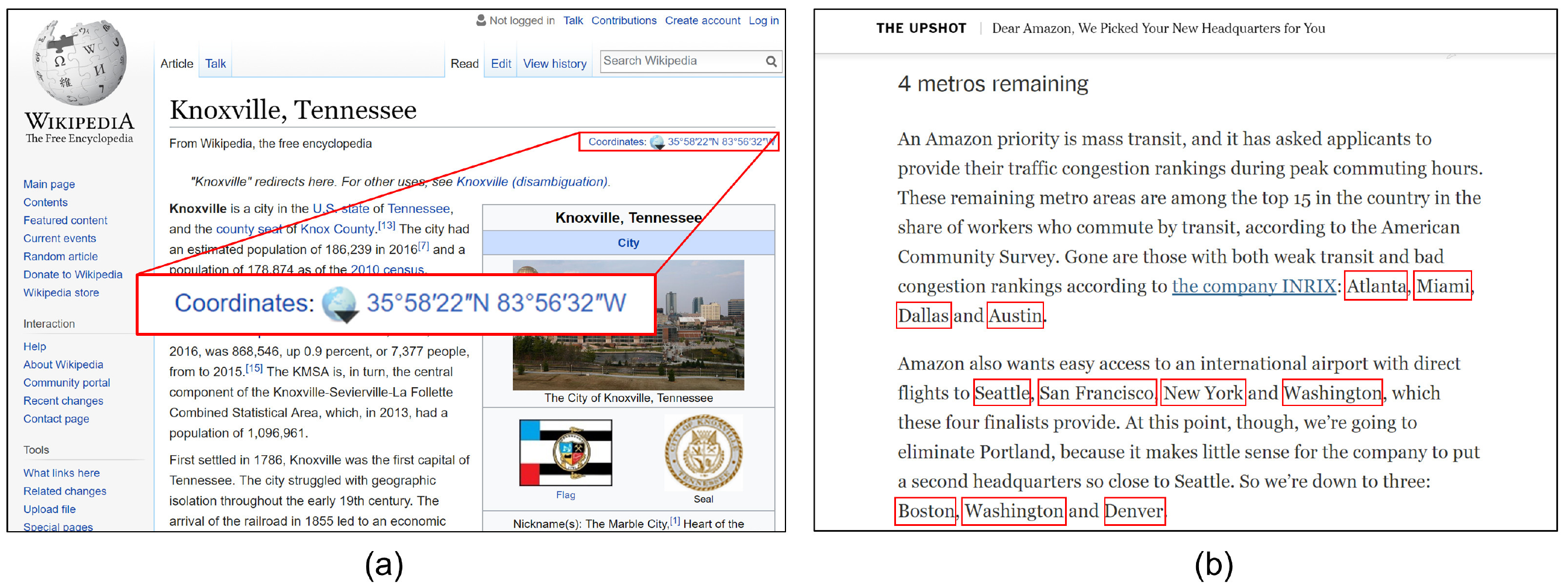}
	\caption{Explicit and implicit geo-text data: (a) a geotagged Wikipedia page; (b) a news article containing city names.}
	\label{Figure_implicit_explicit}
\end{figure}

Depending on how locations are linked to texts, we can differentiate \textit{explicit} and \textit{implicit} geo-text data. \textit{Explicit} geo-text data have spatial footprints explicitly attached to texts, such as geotagged Wikipedia articles (Figure \ref{Figure_implicit_explicit}(a)), while \textit{implicit} geo-text data do not have explicit spatial footprints, but mention place names in their texts, such as the news article in Figure \ref{Figure_implicit_explicit}(b). Implicit geo-text data can be converted to explicit data through geoparsing \citep{gregory2015geoparsing}.



\begin{figure}[h]
	\centering
	\includegraphics[width=0.9\textwidth]{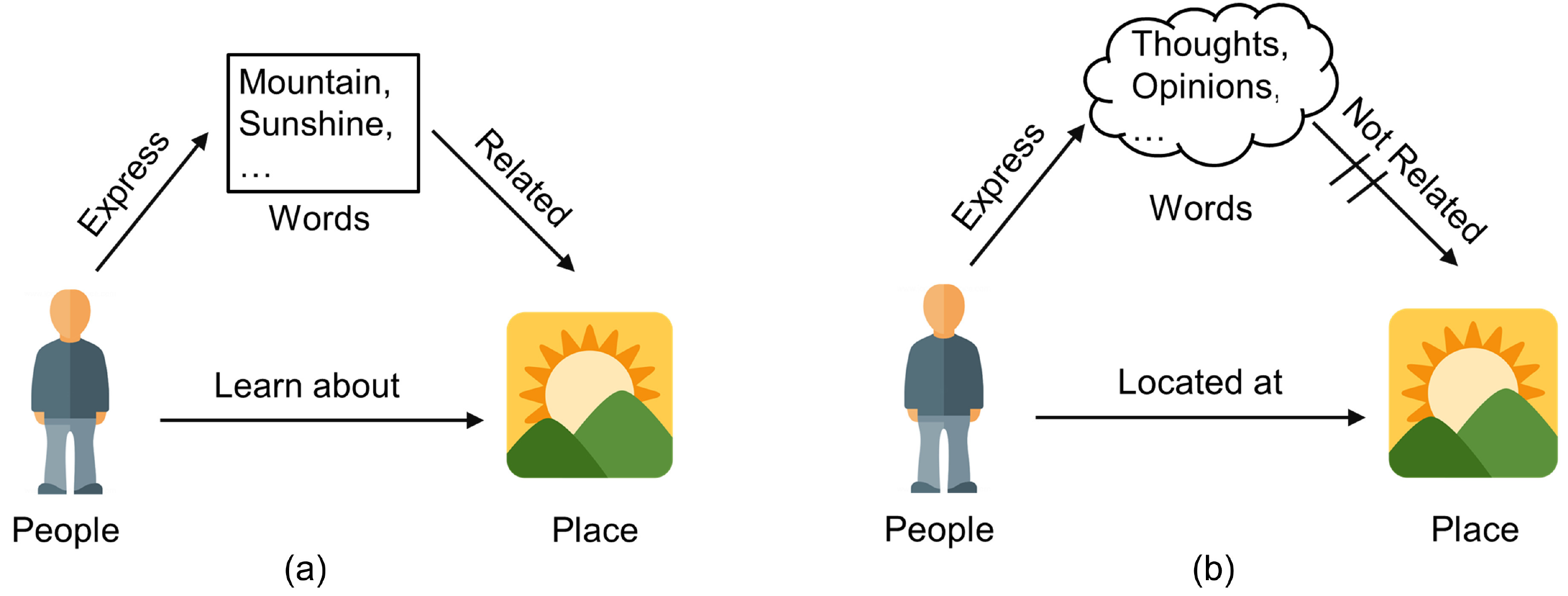}
	\caption{Two major processes that generate geo-text data: (a) people learn about a place and express words related to this place; (b) people are located at a place and express words not directly related to this place.}
	\label{Figure_geo-text_generate}
\end{figure}

Two major processes can generate geo-text data, each of which can generate both explicit and implicit geo-text data. In the first process, people learn about a place and express words related to it (Figure \ref{Figure_geo-text_generate}(a)). For example, an Instagram user may take a photo at a location and describe what he sees (which generates explicit geo-text data); or a travel blog writer may write her experience after a day of touring in a foreign city (which generates implicit data).
In the second process, people  express thoughts and opinions which are not directly related to their current locations (Figure \ref{Figure_geo-text_generate}(b)). For example, a Twitter user may post a tweet irrelevant to his location (explicit geo-text data are generated); or a journalist in her office at Washington D.C. may write a news article about an event happened in Las Vegas (implicit data are generated). The two processes also suggest that the text in a geo-text data record can be linked to two locations: one \textit{about} location and one \textit{from} location. These two locations can be the same, partially overlap, or completely different. Depending on the application needs, we may choose one location over the other or use both simultaneously. In a study on geotagged Twitter data, \mbox{\cite{maceachren2011senseplace2}} made a distinction between tweets \textit{about} and \textit{from} locations. The two processes discussed here are similar to their distinction, but are generalized to geo-text data beyond only geotagged tweets. 

In summary, this section addresses the question ``\textit{what is geo-text data?}" by providing a formal representation and discussing related concepts. Geo-text data can be considered as a collective term that encompasses various types of data that contain links between locations and texts. This section discusses the spatial footprints, timestamps, and texts of geo-text data, and compare it with typical GIS data.  Explicit and implicit geo-text data are differentiated, and  two major processes that generate geo-text data are discussed. 

\section{Knowledge discovery from geo-text data and data-driven geospatial semantics}
The  large volume and rich variety of geo-text data enable the discovery of knowledge that can support disaster response, urban planning, transportation management, and many other applications. Particularly, geo-text data facilitates research in \textit{data-driven geospatial semantics}, a bottom-up approach for studying the meaning of geographic features and terms. \textit{Data-driven geospatial semantics} can be distinguished from the \textit{expert-driven} or top-down approach \citep{kuhn2005geospatial,Hu201880}. Consider measuring the semantic similarity between two words, ``road" and ``street". We can take an \textit{expert-driven} approach by inviting a group of experts to assign scores between 0 and 1 and then taking an average. Alternatively, we can use a \textit{data-driven} approach by harvesting millions of Web pages that contain either ``road" or ``street" and measuring their semantic similarity using context words. Both approaches have their pros and cons, and can sometimes be combined \citep{hu2016enriching}. This section focuses on data-driven research. It reviews a large number of studies and organizes them based on the types of knowledge discovered from geo-text data.



\subsection{Place names}


Extracting place names, or \textit{toponyms}, from texts is a topic frequently studied in GIR \citep{jones2008geographical,wing2011simple,vasardani2013locating,gelernter2013algorithm,doi:10.1080/17538947.2012.674561,laurini2015geographic,nesi2016geographical}. This process is often referred to as \textit{geoparsing} which involves two main steps: toponym recognition and toponym resolution. In toponym recognition, the goal is to identify the words and phrases  that can represent place names. Gazetteers  \citep{lieberman2011multifaceted,zhang2014geocoding} and linguistic features \citep{freire2011metadata,inkpen2015location} are often utilized for this step. In toponym resolution, the goal is to disambiguate and geo-locate the identified place names. Place name disambiguation is necessary due to both geo/geo ambiguity (i.e., the same term, such as \textit{London}, can refer to different places) and geo/non-geo ambiguity (i.e., the same term, such as \textit{Washington}, can refer to both places and persons)  \citep{amitay2004web,leidner2008toponym}. Methods, such as co-occurrences \citep{overell2008using}, conceptual density \citep{buscaldi2008conceptual}, and topic modeling \citep{ju2016things}, were proposed for place name disambiguation. 

Geo-text data can be used for identifying the spatial footprints of place names as well. \cite{kessler2009bottom} studied vague place names, such as ``Soho", using geotagged photo data from Flickr, Panoramio and Picasa, and developed a clustering method based on Delaunay triangulation to construct their spatial footprints.  \cite{hollenstein2010exploring} and \cite{li2012constructing} used geotagged Flickr photos to derive the spatial footprints of city  names using kernel density estimation (KDE).
 %
\cite{jones2008modelling} used a search engine to harvest Web pages that contain a target vague place name (e.g., ``Mid-Wales"), extracted and geo-located the place names contained in these Web pages, and delineated the spatial footprint of the target place name using KDE. 
Table \ref{Table_Place_name} summarizes the studies discussed above. Good reviews focusing on the topic of geoparsing and GIR can also be found in \cite{monteiro2016survey}, \cite{melo2017automated}, and \cite{purves2018geographic}.
\begin{table}[h]
	\centering
	\caption{Summary of the discussed studies on extracting place names and their spatial footprints.}
	\label{Table_Place_name}
	\footnotesize
	\begin{tabular}{|p{5.3cm}|p{4.9cm}|p{4.9cm}|}
		\hline
		Study                                                 & Main Task                                                                                       & Methods                                                                                    \\ \hline

		\begin{tabular}[c]{@{}l@{}}\cite{lieberman2011multifaceted}\\ 
		\cite{freire2011metadata}\\
		\cite{zhang2014geocoding}\\	
		 \cite{inkpen2015location}\end{tabular}                                              & \begin{tabular}[c]{@{}l@{}}Place name recognition \\ and resolution\end{tabular}              & \begin{tabular}[c]{@{}l@{}}Digital gazetteers;
		 \\ Linguistic features; \\
		 Machine learning models\\  \end{tabular}          \\ \hline

		\begin{tabular}[c]{@{}l@{}} \cite{overell2008using}\\ \cite{buscaldi2008conceptual}\\ \cite{ju2016things} \end{tabular}                          & Place name disambiguation                                                                     & \begin{tabular}[c]{@{}l@{}}Co-occurrence models;\\ Conceptual density;\\ Topic modeling\end{tabular} \\ \hline

		\begin{tabular}[c]{@{}l@{}}\cite{kessler2009bottom}\\  \cite{hollenstein2010exploring} \\ \cite{li2012constructing}\\
		\cite{jones2008modelling}\\\end{tabular}                                              & \begin{tabular}[c]{@{}l@{}}Delineating spatial footprints \\of (vague) place  names\end{tabular}              & \begin{tabular}[c]{@{}l@{}}Kernel density estimation;\\ Delaunay triangulation\end{tabular}          \\ \hline
		
	
	\end{tabular}
\end{table}
\normalsize

\begin{figure}[h]
	\centering
	\includegraphics[width=0.95\textwidth]{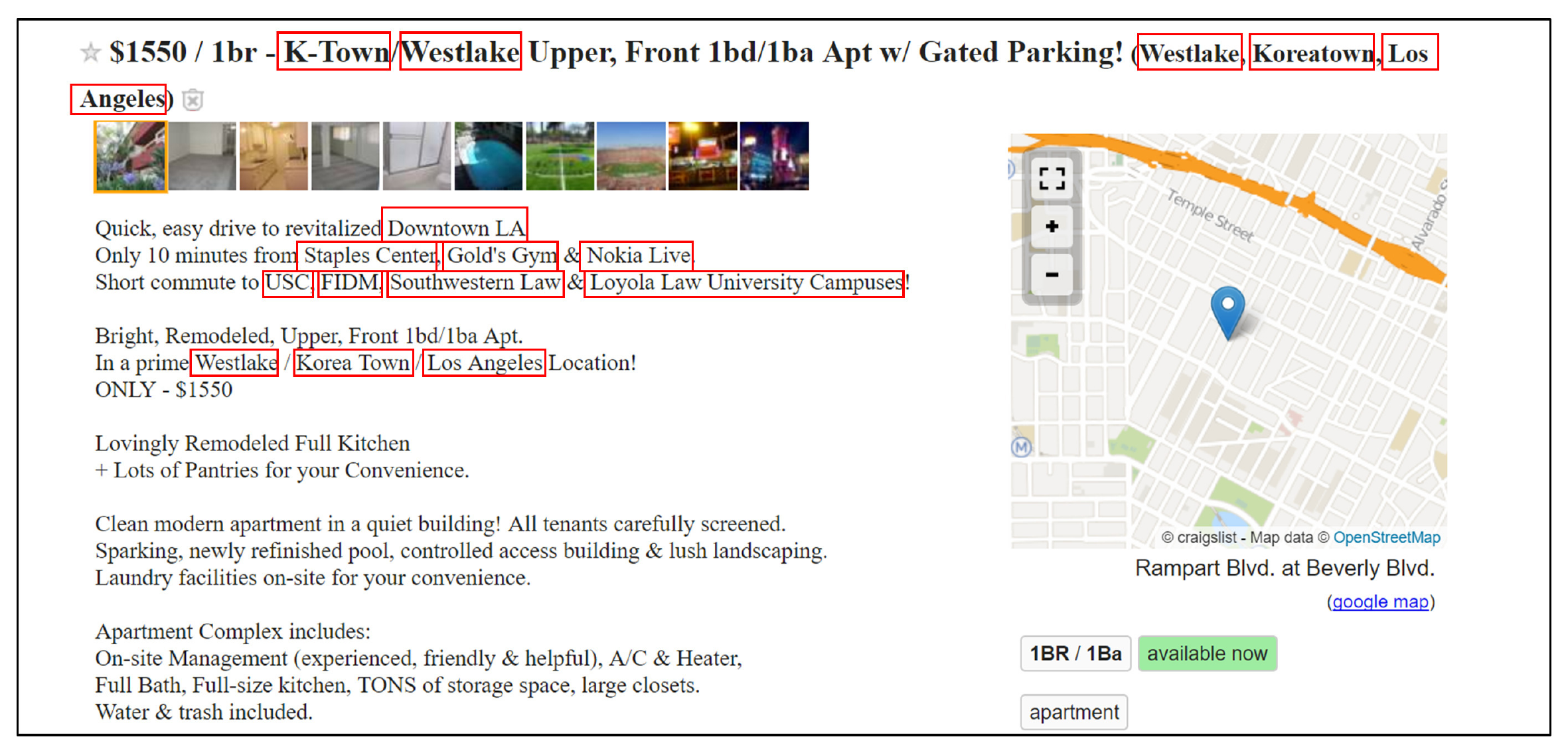}
	\caption{A geotagged housing post published on a local-oriented website in Los Angeles, USA.}
	\label{Figure_ad_vertise}
\end{figure}

Geotagged housing posts are a type of geo-text data that have been rarely studied. They often contain local place names that can be extracted for enriching gazetteers. Figure \ref{Figure_ad_vertise} shows a geotagged housing post published on a local-oriented website. Such a post contains local place names, such as ``K-Town" and ``USC", and the location of the advertised property. One can design methods to extract these local place names and their spatial footprints from these geotagged housing posts.

\subsection{Place relations/sequences}
\begin{figure}[h]
	\centering
	\includegraphics[width=0.95\textwidth]{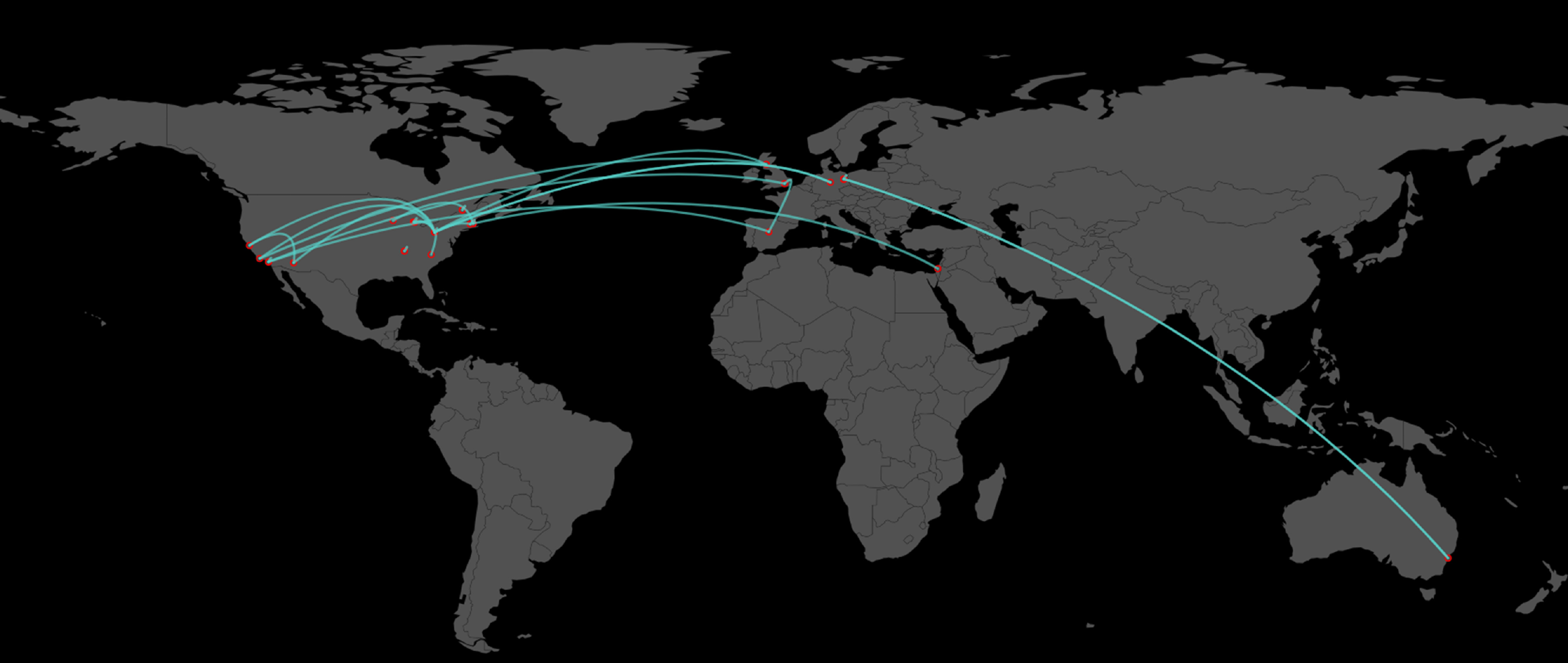}
	\caption{A simple visualization for place relations.}
	\label{Figure4_place_relation}
\end{figure}

The ability of extracting place names from texts enables further examinations on place relations/sequences based on geo-text data. Such examinations can reveal interesting  and sometimes intangible connections among places. Two places can be considered as related if they co-occur in texts or if there exists Web links between them  \citep{ballatore2014evaluative,liu2014analyzing,spitz2016so}. \cite{hecht2009terabytes} conducted an early study using the hyperlinks among geotagged Wikipedia pages to empirically verify Tobler's First Law. They found that nearby places are indeed more likely to have relations than distant ones, although places far away may still have relations. \cite{salvini2016spatialization} analyzed place name co-occurrences in Wikipedia pages, and used the categories of Wikipedia pages to annotate the semantics of place relations. \cite{adams2016exploratory} performed chronotopic analysis on Wikipedia corpus by analyzing the co-occurrences of places and times in texts.  Using news articles, \cite{liu2014analyzing} examined place name co-occurrences and found that place relatedness in news articles decreases less rapidly with the increase of distance, compared with the results from  human movement analysis.  \cite{hu2017extracting} took a topic modeling approach to understand the semantics of place relations using news articles, and found that geographic distance has a non-uniform impact on place relatedness under different topics. City network research also analyzed place name co-occurrences in news articles often based on small data samples and using manual content analysis \citep{taylor1997hierarchical,beaverstock2000globalization}.  Place relations can be visualized on maps (Figure \ref{Figure4_place_relation} shows a simple example), and  further analysis can be performed based on  these relations. 
 
Place sequences can also be extracted from geo-text data. For example, travel trajectories can be extracted from travel blogs, while life trajectories can be derived from biographic documents, such as one's born, study, marriage, and work places. A related research is from \cite{kessler2012spatial} who analyzed the academic trajectories of GIScience scholars based on their affiliation changes in publications. 
Table \ref{table_place_relation} summarizes the studies discussed above. 
\begin{table}[h]
	\centering
	\caption{Summary of the discussed studies on extracting place relations and sequences.}
	\label{table_place_relation}
	\footnotesize
	\begin{tabular}{|p{5.3cm}|p{4.9cm}|p{4.9cm}|}
		\hline
		Study                                                                                                                       & Main Task                    & Methods                                                                                \\ \hline
		\begin{tabular}[c]{@{}l@{}}\cite{hecht2009terabytes}\\  \cite{salvini2016spatialization}\\  \cite{adams2016exploratory}\end{tabular} & \begin{tabular}[c]{@{}l@{}} Exploring and quantifying place \\and time relations \end{tabular} & \begin{tabular}[c]{@{}l@{}}Place co-occurrences or \\ hyperlinks in Wikipedia pages\end{tabular} \\ \hline
		
		\begin{tabular}[c]{@{}l@{}} \cite{liu2014analyzing}\\  \cite{hu2017extracting}\end{tabular}                                               &
		\begin{tabular}[c]{@{}l@{}} Extracting place relations and \\analyzing distance   decay effects \end{tabular}
		& 
		\begin{tabular}[c]{@{}l@{}}Place co-occurrences in news \\articles and gravity models
	 \end{tabular}                 \\ \hline
 
 	\begin{tabular}[c]{@{}l@{}} \cite{taylor1997hierarchical}\\  \cite{beaverstock2000globalization}\end{tabular}                                               &
 \begin{tabular}[c]{@{}l@{}} Analyzing city relations and \\ city networks \end{tabular}
 & 
 \begin{tabular}[c]{@{}l@{}}Content analysis based on \\sampled newspapers in cities
 \end{tabular}                 \\ \hline
 
		 \cite{kessler2012spatial}                                                                                                        & \begin{tabular}[c]{@{}l@{}} Extracting place sequences \\ and trajectories  \end{tabular}            & \begin{tabular}[c]{@{}l@{}}Author affiliation analysis \\ based on publication records \end{tabular}    \\ \hline
	\end{tabular}
\end{table}
\normalsize

\subsection{Place opinions/emotions}
Geo-text data contains words expressed by people, which make it less suitable for studying environmental variables but more appropriate for examining the opinions and emotions of people. Sentiment analysis is a subfield in natural language processing (NLP) \citep{pang2008opinion,liu2012sentiment}, which 
 also attracted the interests of GIScience researchers. 
Using travel blog data, \cite{ballatore2015extracting} analyzed the emotions of place types 
and constructed a vocabulary that associates  sentiment words with place types.  Based on geotagged tweets, \cite{nelson2015geovisual} developed a geovisual analytics tool, called \textit{SPoTvis}, and applied it to the tweets related to the debate on the Affordable Care Act in the US. 
\cite{wang2016spatial} performed text mining on geotagged tweets in the response to a wildfire, and detected the attitudes of people, such as their appreciation to fire fighters. 
Looking into TripAdvisor hotel reviews, \cite{cataldi2013good} proposed an approach for detecting the sentiments of people toward different aspects of a hotel, such as its location convenience and food quality. \cite{wang2016geography} performed sentiment analysis on TripAdvisor hotel reviews  within the same city, and found that spatial dependence exists in the satisfaction of customers.  \cite{wartmann2018investigating} investigated the sense of place of people towards landscape features (e.g., mountains and rivers) by collecting free listings and place descriptions from visitors, and found that the elicited sense of place was similar across landscape types. Table \ref{table_place_emotions} summarizes the discussed studies.
\begin{table}[h]
	\centering
	\caption{Summary of the discussed studies on extracting place opinions and emotions.}
	\label{table_place_emotions}
	\footnotesize
	\begin{tabular}{|p{5.3cm}|p{4.9cm}|p{4.9cm}|}
		\hline
		Study                                                                               & Main Task                                                                 &  Methods                                                                                  \\ \hline
		 \cite{ballatore2015extracting}                                                          & \begin{tabular}[c]{@{}l@{}} Extracting the emotions\\ associated with place types\end{tabular}                                                & \begin{tabular}[c]{@{}l@{}}Sentiment analysis based on \\travel blog data\end{tabular}        \\ \hline
		 
		  \cite{nelson2015geovisual}                                                                & \begin{tabular}[c]{@{}l@{}}Visualizing and analyzing \\opinions on political events\end{tabular} & \begin{tabular}[c]{@{}l@{}}Geovisual analytics based on \\geotagged tweets\end{tabular}       \\ \hline
		 
	 \cite{wang2016spatial}                                                                  & \begin{tabular}[c]{@{}l@{}}Understanding the attitudes of \\ people in disaster response  \end{tabular}                                                  & \begin{tabular}[c]{@{}l@{}}Text analysis based on  \\ geotagged tweets\end{tabular}            \\ \hline
		
		\begin{tabular}[c]{@{}l@{}} \cite{cataldi2013good}\\  \cite{wang2016geography}\end{tabular} & \begin{tabular}[c]{@{}l@{}}Examining the sentiments of \\people toward hotels\end{tabular}                                                    & \begin{tabular}[c]{@{}l@{}}Sentiment analysis based on \\TripAdvisor hotel reviews  \end{tabular} \\ \hline

	 \cite{wartmann2018investigating}                                                                  & \begin{tabular}[c]{@{}l@{}}Investigating sense of place \\ towards landscape features  \end{tabular}                                                  & \begin{tabular}[c]{@{}l@{}}Interviews, frequent terms, and \\  similarity comparisons \end{tabular}            \\ \hline
		
	\end{tabular}
\normalsize
\end{table}

\begin{figure}[h]
	\centering
	\includegraphics[width=\textwidth]{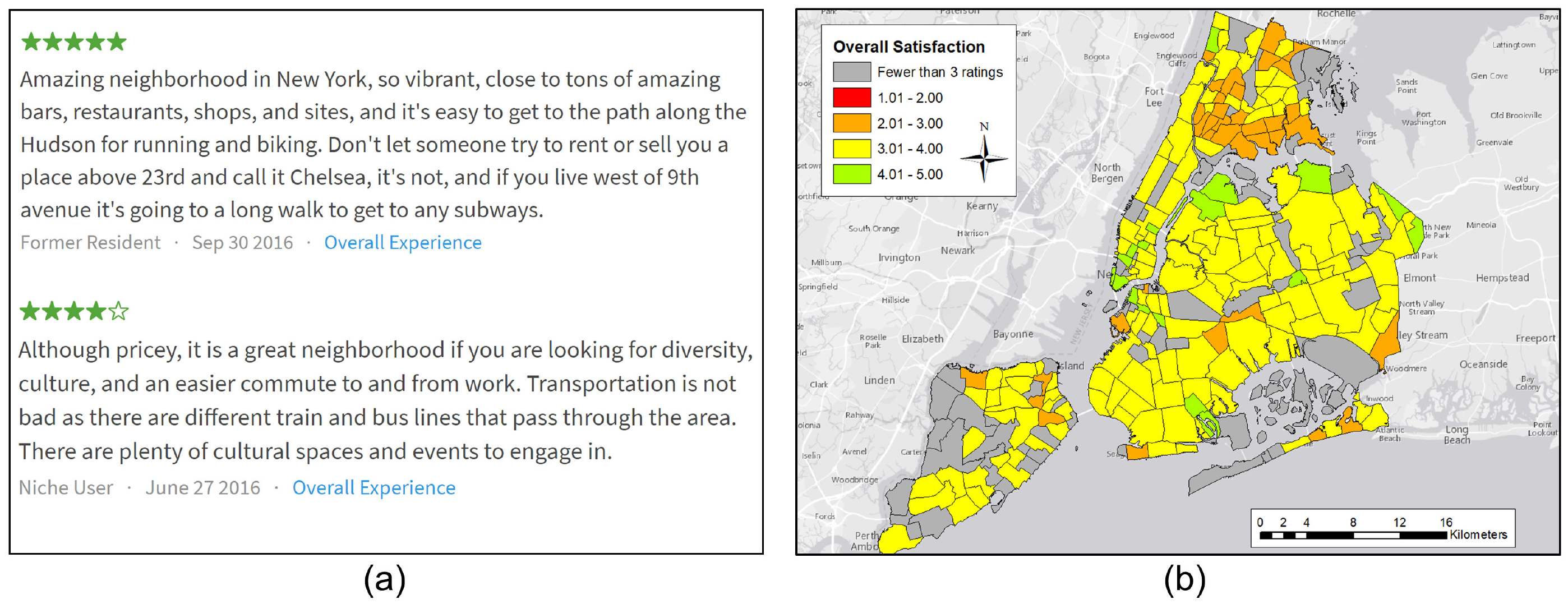}
	\caption{(a) Two reviews  from a neighborhood review website; (b) the overall satisfaction levels of people toward neighborhoods (the highly-satisfied are in green while the less-satisfied are in orange).}
	\label{Figure5_place_emotion}
\end{figure}

Neighborhood reviews are a relatively new type of geo-text data. In recent years, there is an emergence of websites, e.g., StreetAdvisor and Niche, designed to help people find suitable neighborhoods to live. On these websites, current or previous residents can review their neighborhoods. Figure \ref{Figure5_place_emotion}(a) shows two reviews on a neighborhood in New York City (NYC), and Figure \ref{Figure5_place_emotion}(b) shows the overall satisfaction levels  based on the review ratings. Analyzing these reviews can help understand the perceptions of people, and can benefit urban planning and quality of life studies  \citep{kessler2005argumentation,das2008urban}.  


\subsection{Place zones}
\begin{figure}[h]
	\centering
	\includegraphics[width=\textwidth]{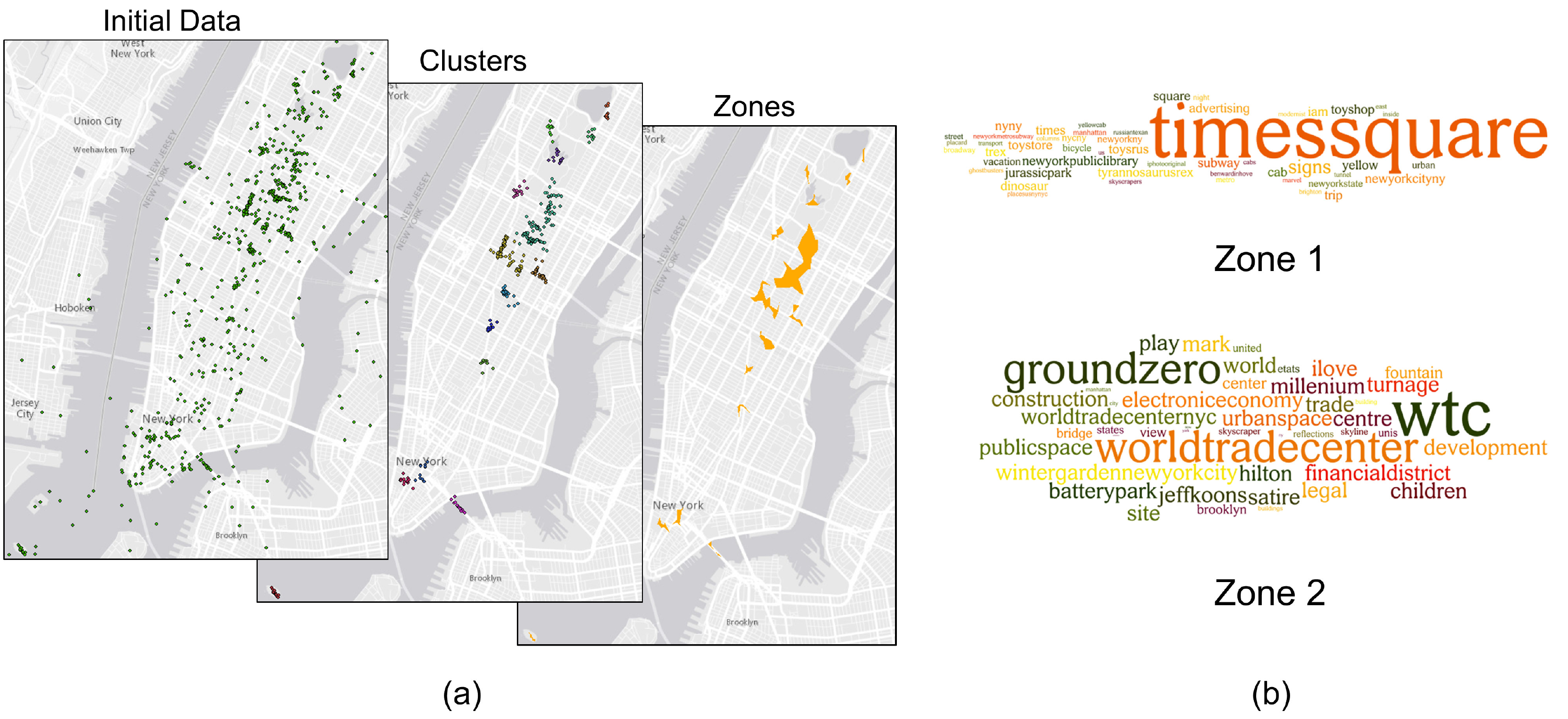}
	\caption{(a) Place zones extracted from geotagged Flickr photo data in NYC; (b) representative words of two zones. }
	\label{Figure6_place_zone}
\end{figure}

Place zones are another type of knowledge that can be extracted from geo-text data. While detecting hot zones from location data is a common GIS operation,
the uniqueness of  using geo-text data  lies in its rich semantics: we can understand the diverse reasons underlying the formation of these zones based on the words of people.
There exist many empirical studies on extracting place zones using geo-text data. Based on geotagged Flickr photos, \cite{rattenbury2009methods} identified point clusters using K-means clustering, and detected representative textual tags for each cluster using an algorithm called TagMaps. \cite{andrienko2010discovering} proposed a visual analytics framework which detects special place zones containing periodic or irregular events. 
\cite{hu2015extracting} used geotagged Flickr photos  to extract urban areas of interest (AOI) by performing DBSCAN clustering and chi-shape algorithm (Figure \ref{Figure6_place_zone}(a)). They employed term frequency and inverse document frequency (TF-IDF) to identify representative words for the extracted AOI (Figure \ref{Figure6_place_zone}(b)). 
\cite{velasco2017secrets} identified place zones in the city of Quito, Ecuador based on venue locations and textual reviews on TripAdvisor, and labeled these zones with user-generated texts. Some research combined multiple types of data. For example, 
\cite{jenkins2016crowdsourcing} examined both geotagged tweets and Wikipedia pages, and applied topic modeling, semantic analysis, and geospatial clustering to find locations with a collective sense of place. \cite{gao2017data} synthesized geotagged social media data from Twitter, Flickr, and Instagram  to extract cognitive zones such as ``SoCal" and ``NorCal". Using the data from Foursquare, Twitter, and Yik Yak, \cite{mckenzie2017juxtaposing} compared the regions identified based on place instances and place mentions. 
Table \ref{table_placezones} summarizes the discussed studies.
\begin{table}[h]
	\centering
	\caption{Summary of the discussed studies on extracting place zones  from geo-text data.}
	\label{table_placezones}
	\footnotesize
	\begin{tabular}{|p{5.3cm}|p{4.9cm}|p{4.9cm}|}
		\hline
		Study                        & Main Task                                                                                            &  Methods                                                                                            \\ \hline
		
		\begin{tabular}[c]{@{}l@{}}  \cite{rattenbury2009methods} \end{tabular}
		& \begin{tabular}[c]{@{}l@{}}Extracting representative tags\\ for special zones within a city\end{tabular}     & \begin{tabular}[c]{@{}l@{}}K-means clustering and \\TagMaps on geotagged photos\end{tabular}    \\ \hline
		
		\cite{andrienko2010discovering}     & \begin{tabular}[c]{@{}l@{}}Detecting special place areas\\ that  contain events\end{tabular}                  & \begin{tabular}[c]{@{}l@{}}A visual analytics framework \\with event  detection\end{tabular}             \\ \hline
		
		 \cite{hu2015extracting}             & \begin{tabular}[c]{@{}l@{}}Extracting urban AOI and \\representative words\end{tabular}                      & \begin{tabular}[c]{@{}l@{}}DBSCAN clustering, chi-shape \\algorithm, and TF-IDF \end{tabular}      \\ \hline
		 
		  \cite{velasco2017secrets}        & \begin{tabular}[c]{@{}l@{}} Identifying place zones in the \\city of Quito, Ecuador\end{tabular}                & \begin{tabular}[c]{@{}l@{}}K-means clustering and \\ TF-IDF  on TripAdvisor data\end{tabular}      \\ \hline

	 \cite{jenkins2016crowdsourcing}        & \begin{tabular}[c]{@{}l@{}}Identifying locations with a \\collective sense of place\end{tabular}               & \begin{tabular}[c]{@{}l@{}}Topic modeling, semantic \\ analysis,  and spatial clustering\end{tabular} \\ \hline

     \cite{gao2017data}           & \begin{tabular}[c]{@{}l@{}}Extracting vague cognitive \\ zones and semantic topics\end{tabular}                        & \begin{tabular}[c]{@{}l@{}} Clustering and topic modeling \\ based on multiple types of data \end{tabular} \\ \hline

 	\begin{tabular}[c]{@{}l@{}}\cite{mckenzie2017juxtaposing} \end{tabular}
 	& \begin{tabular}[c]{@{}l@{}}Comparing the regions \\extracted in different ways \end{tabular} & \begin{tabular}[c]{@{}l@{}}KDE on multiple types of\\ social media data\end{tabular}           \\ \hline
	
	\end{tabular}
\end{table}
\normalsize

\subsection{Place impacts}
Geo-text data can help reveal the impact of an event, such as a natural disaster, a public policy, an infectious disease, or others (all are referred to as \textit{target event}). From texts, we can understand the attitudes  of people; from locations, we can examine the geographic areas where people are reacting to the event. Here, we  focus more on the \textit{from} location rather than the \textit{about} location of geo-text data. 
 In addition, the impacts examined here are \textit{social} rather than \textit{physical} impacts of events.

Many studies have utilized geo-text data to investigate the impacts of events. A notable example is Google's Flu Trends (GFT) \citep{ginsberg2009detecting}, in which the search keywords from people were linked to their locations based on IP addresses to predict the intensities of influenza-like illness (ILI) in different geographic areas. While GFT eventually failed, it nevertheless demonstrated a novel idea by linking  texts to locations. Based on a sample of news articles, \cite{wang2015spatiotemporal} examined the impact of Hurricane Sandy by extracting place names, timestamps, and emergency information (e.g., power failure). Also using news articles,  \cite{peuquet2015method} developed a computational method that extends the T-pattern analysis, and applied this technique to discovering the event associations during the Arab Spring.
Geotagged tweets are widely used for studying place impacts \citep{tsou2015research,haworth2015review}. Focusing on public health issues, \cite{issa2017understanding} studied the spatial diffusion of tweets about flu in four different cities, while \cite{nagar2014case} used daily geotagged tweets in NYC to investigate the spatiotemporal tweeting behavior related to ILI. Looking into disaster responses, \cite{crooks2013earthquake} examined the spatial and temporal characteristics of tweets after an earthquake, while \cite{de2009omg} investigated the tweeting activities during a major forest fire. There exist other methods and visual analytics systems for detecting anomalies  \mbox{\citep{chae2012spatiotemporal,thom2012spatiotemporal,andrienko2013thematic}}, extracting topics and events  \mbox{\citep{cho2016vairoma}}, and analyzing place-time-attribute information \citep{pezanowski2017senseplace3}. Table \mbox{\ref{table_place_impact}} summarizes the discussed studies.

\begin{table}[h]
	\centering
	\caption{Summary of the discussed studies on examining place impacts based on geo-text data.}
	\label{table_place_impact}
	\footnotesize
	\begin{tabular}{|p{5.3cm}|p{4.9cm}|p{4.9cm}|}
		\hline
		Study                                                                                                                                     & Main Task                                                                                    &  Methods                                                                                              \\ \hline
	 \cite{ginsberg2009detecting}                                                                                                                    & \begin{tabular}[c]{@{}l@{}}Predicting the locations and \\intensities of ILI\end{tabular}   & \begin{tabular}[c]{@{}l@{}}Examining the spatiotemporal \\patterns of search keywords\end{tabular}            \\ \hline
	 \cite{wang2015spatiotemporal}                                                                                                                   & \begin{tabular}[c]{@{}l@{}}Understanding the  impacts\\ of Hurricane Sandy\end{tabular}       & \begin{tabular}[c]{@{}l@{}}Extracting place and event \\ information from news articles\end{tabular}              \\ \hline
	 \cite{peuquet2015method}                                                                                                                     & \begin{tabular}[c]{@{}l@{}}Discovering the  associations \\of social and political events\end{tabular}          & \begin{tabular}[c]{@{}l@{}}A computational method that \\extends the T-pattern  analysis\end{tabular}                       \\ \hline
		\begin{tabular}[c]{@{}l@{}} \cite{issa2017understanding}\\  \cite{nagar2014case}\\  \cite{crooks2013earthquake}\\  \cite{de2009omg} \end{tabular} & \begin{tabular}[c]{@{}l@{}}Understanding the spatial  \\and temporal impacts of \\diseases and disasters\end{tabular}  & \begin{tabular}[c]{@{}l@{}}Spatial, temporal, and content\\ analysis on geotagged tweets\end{tabular}    \\ \hline
		

	\end{tabular}
\end{table}
\normalsize

\begin{figure}[h]
	\centering
	\includegraphics[width=\textwidth]{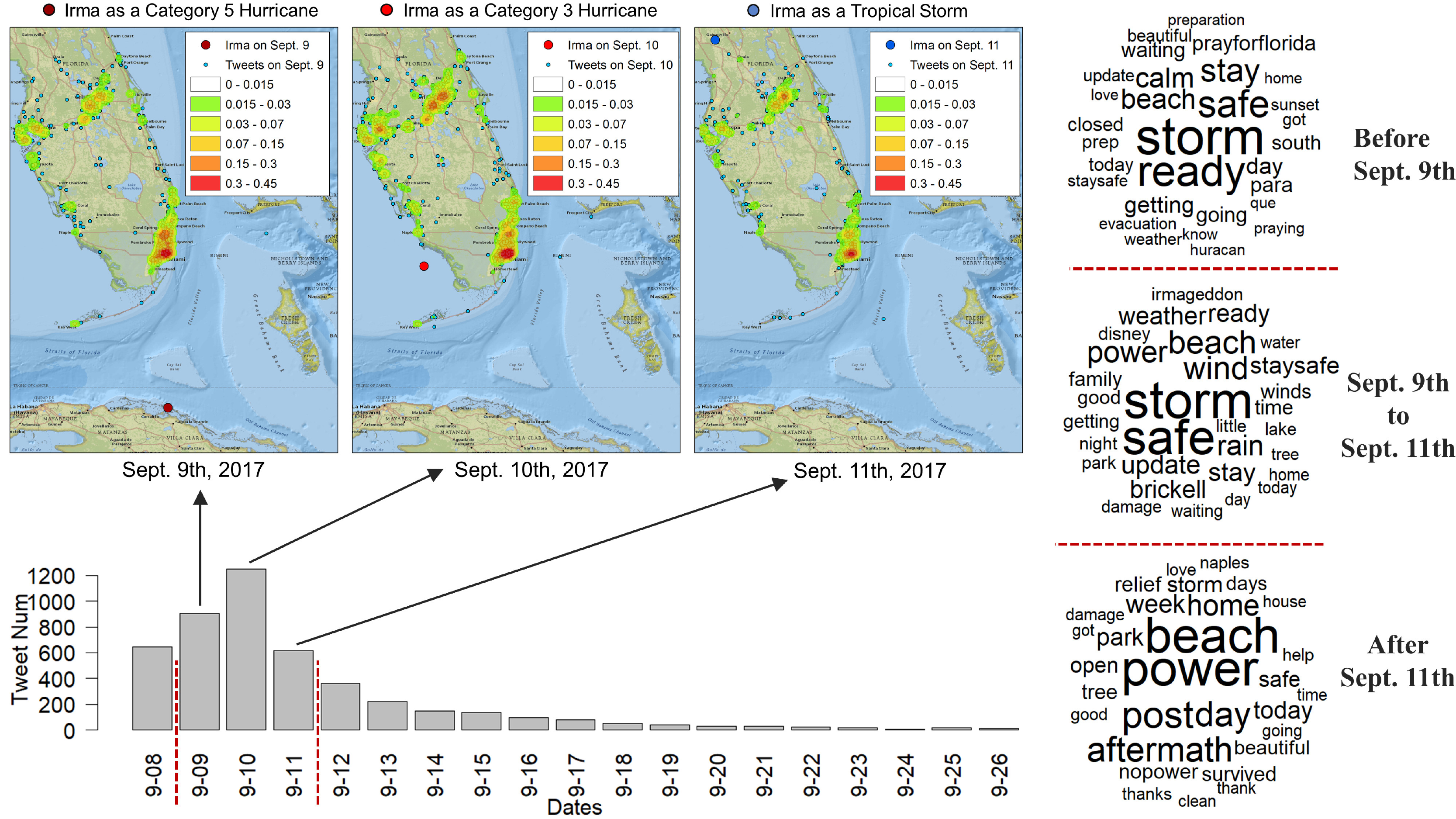}
	\caption{Exploratory analysis on the impact of Hurricane Irma based on a sample of geotagged tweets.}
	\label{Figure7_place_impact}
\end{figure}

With its ability of capturing real-time public reactions, geotagged tweets have become a convenient resource for exploring the impact of an event.  One simple approach is to first retrieve related tweets using keywords, and then examine the spatiotemporal patterns of the retrieved tweets. Figure \ref{Figure7_place_impact} shows an example of using this  approach for exploring the impact of Hurricane Irma in September 2017. From the tweet counts on different days (at the bottom of the figure), we can see that most tweets were posted between Sept. 9th and 11th when Irma was landing and moving inland Florida. By dividing the entire dataset into three groups (based on the two red dotted lines), we can see  different frequent words at different stages of the hurricane. By visualizing the tweet locations on three specific days (at the top of the figure), we can see the major areas of the tweets. This spatial-temporal-thematic analysis also shows that the place impact examined here is \textit{social} rather than \textit{physical} impact, since most tweets came from the cities rather than the areas lying on the hurricane path.

%

\subsection{Summary}
This section answers the question ``\textit{what can we get from geo-text data?}" by reviewing a large number of studies, discussing their used datasets and methods, and organizing them based on the types of knowledge discovered. Place names are important geographic information that can be extracted from texts, and are necessary for more advanced tasks, such as place relation or sequence analysis. We can investigate the opinions and sentiments of people attached to places, and can identify place zones and understand their meanings. We can also study the impact of an event by examining the attitudes of people in different geographic areas. While five types of knowledge are identified here, this list is not exhaustive and other types of knowledge can be discovered as well. 

\section{Towards a generalized workflow for analyzing geo-text data}
\begin{figure}[h]
	\centering
	\includegraphics[width=\textwidth]{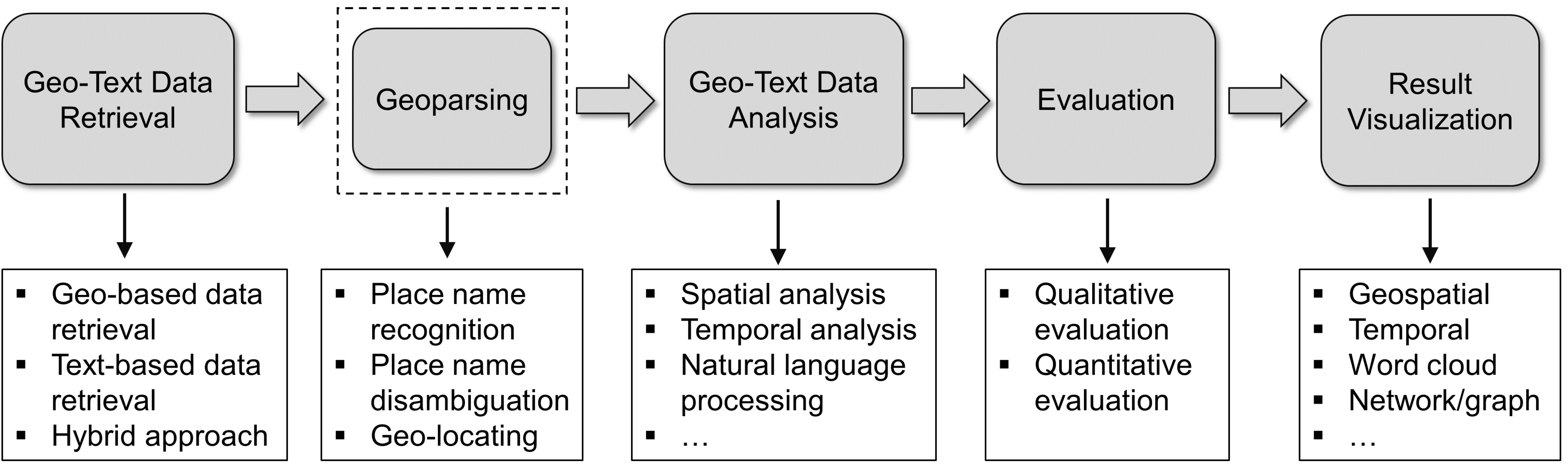}
	\caption{A generalized workflow for analyzing geo-text data.}
	\label{Figure8_workflow}
\end{figure}

While previous studies used different types of geo-text data and examined problems in various domains, they share similar procedures in data processing and analysis. This section extracts a  workflow from the previous studies in order to 
provide a general reference for future research. Figure \ref{Figure8_workflow} illustrates this workflow.

\textbf{Geo-text data retrieval.} Retrieving relevant geo-text data  is the first step for conducting a study. The retrieval process can be performed in three major ways. The first one is based on location, in which a bounding extent is often used. For example, we can
retrieve all geotagged Wikipedia pages within the boundary of California. The second approach is text-based data retrieval, in which keywords or topics are used to retrieve data. For example, we can collect a set of news articles related to certain keywords. The third one is a hybrid approach which retrieves geo-text data using both locations and texts. 


\textbf{Geoparsing.} 
This step recognizes the words that represent place names from texts, resolves the ambiguous names, and geo-locates the place names to their corresponding spatial footprints. This step is within a dotted rectangle in Figure \ref{Figure8_workflow}, since explicit geo-text data may not need  geoparsing. While multiple geoparsers exist, 
they can have very different performances   when applied to different testing corpora \citep{monteiro2016survey}. \cite{gritta2017s} tested five geoparsers using the same datasets, and their performances based on one dataset are provided in Table \ref{geoparser_Performance}. When applied to a corpus with many ambiguous place names, the performance of a geoparser can decrease dramatically  \citep{ju2016things}. 
\begin{table}[h]
	\centering
	\caption{Performances of five  geoparsers applied to the same dataset \mbox{\citep{gritta2017s}}.}
	\label{geoparser_Performance}
	\footnotesize
	\begin{tabular}{|p{6cm}|p{2.3cm}|p{2.3cm}|p{2.3cm}|}
		\hline
		& Precision & Recall & F-score \\ \hline
		GeoTxt \mbox{\citep{karimzadeh2013geotxt}}      & 0.80      & 0.59   & 0.68    \\ \hline
		Edinburgh \mbox{\citep{alex2015adapting}}   & 0.71      & 0.55   & 0.62    \\ \hline
		Yahoo! PlaceSpotter      & 0.64      & 0.55   & 0.59    \\ \hline
		CLAVIN      & 0.81      & 0.44   & 0.57    \\ \hline
		TopoCluster \mbox{\citep{delozier2015gazetteer}} & 0.81      & 0.64   & 0.71    \\ \hline
	\end{tabular}
\end{table}
\normalsize



\begin{figure}[h]
	\centering
	\includegraphics[width=0.67\textwidth]{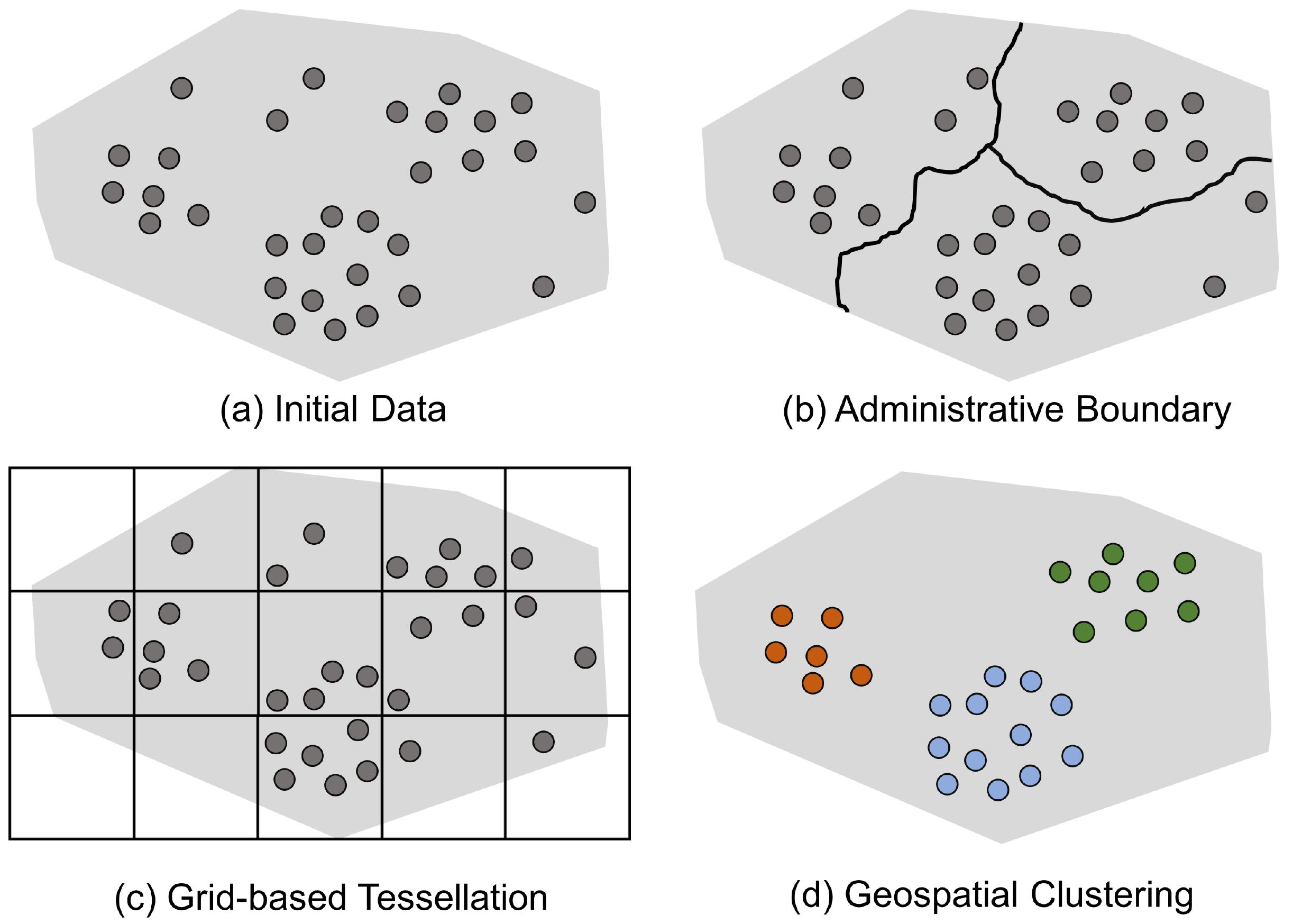}
	\caption{Three methods for grouping geo-text data based on their locations.}
	\label{Figure9_geo_first}
\end{figure}

\textbf{Geo-text data analysis.} This is a key step in the workflow. Many methods can be utilized, such as named entity recognition, topic modeling, and sentiment analysis for the text part, and KDE, geospatial clustering, spatial autocorrelation for the location part. When timestamps are available, temporal analysis can also be performed.  
More systematically, we can categorize the analysis process into \textit{geo-first} and \textit{text-first}. In \textit{geo-first}, we start from the locations of geo-text data by segmenting or grouping them. Figure \ref{Figure9_geo_first} shows three examples of segmenting data using the administrative boundary, a grid-based tessellation, and a clustering technique respectively.
Sometimes, the identified data groups can also spatially overlap. With the grouped data, we can use the texts associated with each group to examine its semantics. The work of \cite{andrienko2010discovering} is an example of the \textit{geo-first} approach. In \textit{text-first}, we begin with the text part  by extracting information from it, and then investigate the spatial or spatiotemporal patterns of the extracted information. For example, we can first extract place relations based on place name co-occurrences in texts, and then explore their patterns in geographic space. The work of \cite{peuquet2015method} is an example of the \textit{text-first} approach. 

\textbf{Evaluation.} Evaluation is critical for ensuring that the extracted knowledge is valid and useful. Evaluations for geo-text data can benefit from both qualitative and quantitative assessments. While qualitative assessment is sometimes criticized for its lack of representativeness, it provides intuitive understandings  on the obtained results. 
Meanwhile, quantitative assessment is necessary for robust evaluations. Suitable quantitative metrics are project specific, but common ones include accuracy, error rate, and correlation coefficient. It is usually better to combine qualitative and quantitative assessments  to provide both intuitive and robust evaluations.

\textbf{Result visualization.} Many techniques can be employed to visualize the knowledge extracted from geo-text data. Given the  locations and timestamps, we can visualize the results as  maps, temporal sequences, or spatiotemporal cubes \citep{andrienko2010space,luo2014geo,nelson2015geovisual}. With  texts, techniques, such as word clouds, multi-dimensional scaling, or self-organization maps,  can be used \citep{skupin2003spatialization}. 
Network or graph visualizations can be employed to show the extracted place relations. 
There also exist geovisual analytics systems, such as SensePlace2 \citep{maceachren2011senseplace2}, SensePlace3 \citep{pezanowski2017senseplace3},  STempo \citep{robinson2017design}, and VAiRoma \citep{cho2016vairoma}, that support overview and detailed visualization of data. 


In summary, this section answers the question ``\textit{what is a general workflow that we can follow to analyze geo-text data?}" by extracting a step-by-step data analysis skeleton from the literature. One can flesh out a specific workflow by choosing a data retrieval approach, selecting a geoparser, deciding data analysis methods, designing evaluation experiments, and choosing visualization techniques. Such a workflow also carries important implications that should be noted. For the retrieved geo-text data, the data source  can directly affect the analysis results. For example, depending on whether the data are generated by tourists or local residents, the identified place zones may reflect the different interests of the two groups of people. For the step of geoparsing, while methods and geoparsers have been developed, they are not perfect and errors  can propagate to the downstream of data analysis. For geo-text data analysis, the chosen methods can have certain assumptions. For example, term frequency analysis assumes that the importance of a term  is reflected in its mentioning frequency. However, it is possible that some important terms are mentioned only a few times. Finally, the selected data visualization methods can also distort the perceptions of readers. These implications do not mean a decreased value of geo-text data analysis. In fact, they make some problems, e.g., local versus tourist places, more interesting. However, these implications should be kept in mind when we interpret the analysis results.


\section{Challenges for future research}
This section discusses some of the key challenges for future research based on geo-text data. The discussion is organized based on the dual parts of geo-text data, the special link between them, and the  development of future methods. 

\textbf{Uncertainty of spatial footprints.} As discussed previously,  the spatial footprints of geo-text data can be points, lines, polygons, and even polyhedra. Accordingly, geo-text data can be affected by the same  uncertainty issues like  other  GIS data \citep{ehlschlaeger1997visualizing}. When the spatial footprint is a vague region, suitable representation methods need to be selected and ideally verified with human perceptions \citep{montello2003s,jones2008modelling}. In addition, some existing geo-text data have only point-based footprints (e.g., the point location of Knoxville in Figure \ref{Figure_implicit_explicit}(a)), whereas the geographic features may be better represented using polylines or polygons  given the application scale. In short, how can we represent the spatial footprint of geo-text data more accurately based on application needs? 

\textbf{Ambiguity of texts.} Despite the advancements in NLP, it is still challenging to  accurately understand natural language texts. The commonly used bag-of-words model ignores word orders and cannot capture the structures of sentences. Other text processing methods, such as parse trees and smoothing windows \citep{mikolov2013distributed,schuster2016enhanced}, as well as the recent  deep neural nets provide more advanced approaches for modeling texts \citep{tang2015document,li2017leveraging}, but often require large amounts of labeled  corpora and can take much longer time to train than traditional models. Even when the state-of-the-art methods are employed, some of  the entities,  topics, and sentiments extracted from geo-text data can still be incorrect. In short, how can we improve the accuracy of understanding texts given reasonable computing and data resources?      


\textbf{Unclear Links between locations and texts.} The link between geography and text makes geo-text data special. However, such links can be unclear and can bring challenges in two areas. First, the text can be related to multiple locations. As discussed in Section 2,  geo-text data  can have both \textit{about} location and \textit{from} location. Besides, some geo-text data, such as news articles, can mention multiple place names in the same textual context. In these situations, a suitable method needs to be selected to link the text with the right geographic reference. Second, the text may link to only part of a referred geographic feature. For example, one may be referring to only the peak of a mountain or the mouth of a river, and in such situations, it can be inappropriate to link the text to the whole  feature. Likewise, it can also be inappropriate to link the text to only part of a geographic feature when the text in fact refers to the whole feature. In short, how can we correctly and accurately link locations and texts for geo-text data?

\textbf{Loose integration between spatial and text analysis.} Because of the dual parts of geo-text data,  many previous studies integrated spatial and text analysis.   Such integrations, however, were usually in a loose manner, which is also reflected in the \textit{geo-first} or \textit{text-first} approaches identified  in Section 4.  While such loose integration can already discover  useful knowledge, methods closely integrating spatial and text analysis may better address problems in text understanding and geoparsing. For example, the geographic context of a person can help interpret the words of this person; meanwhile, the topics that people talk about can help infer their locations. Some research has  examined such a close integration \citep{cocos2017language}. In addition, using geographic knowledge to improve computational models, rather than simply taking  methods from other fields, can help increase the impact of GIScience overall. In short, how can we more closely integrate locations and texts to develop methods with potential cross-domain impacts? 


In summary, this section answers the question ``\textit{what are some key challenges for future research?}" by identifying the difficulties in four areas. The first two areas, spatial footprint uncertainty and text ambiguity, are related to not only geo-text data but also spatial data and  linguistic data in general. Thus, they can benefit from the  advancements in the corresponding fields. The second two areas are related to the special links between locations and texts, which are more unique to geo-text data and need integrated thinking rather than separated studies. These challenges  provide great opportunities for future research.
 


\section{Conclusions and summary}
\textit{Geo-text data} is a collective term referring to the kind of data that contains  links between geographic locations and natural language texts. Compared with typical GIS data, such as temperature measurements and DEM, in which numeric values are attached to locations, geo-text data links unstructured texts to locations. Such special links make geo-text data unique, and enable new research topics on place names, place relatedness, sense of place, vague spatial footprints, cognitive regions, the attitudes of people toward events, and many other topics that require an additional dimension of human experience.  Geo-text data greatly facilitates research in data-driven geospatial semantics by  enhancing our understanding on the semantics of geographic features and terms. 
This paper provides a formal representation  of geo-text data based on a general GIS theory. Explicit and implicit geo-text data are differentiated, and  two major processes that generate geo-text data are discussed.  A systematic literature review is conducted which organizes  previous studies based on the  types of knowledge  discovered. A generalized workflow is then extracted from the literature, and key challenges  for future research are discussed.
Overall, geo-text data and the related research are situated at the intersection of geography, computer science, linguistics, cognitive science, statistics, and other related fields. Interdisciplinary collaboration is and will continue playing an important role in fostering advancements in this growing and exciting area. 



\section*{Acknowledgements}
The author would like to thank Dr. James Cheshire, Dr. Michael Goodchild, and the anonymous reviewers for their constructive comments and suggestions. This work is supported by the Professional and Scholarly Development Award (Award Number: R011038-002) from the University of Tennessee.


 \bibliographystyle{apalike}

\bibliography{references}

\end{document}